\documentclass[cameraready]{Interspeech}

\title{POTSA: A Cross-Lingual Speech Alignment Framework for Speech-to-Text Translation}

\author[affiliation={1}]{Xuanchen}{Li}
\author[affiliation={1}]{Chenrui}{Cui}
\author[affiliation={1,2}]{Tianrui}{Wang}
\author[affiliation={1}]{Meng}{Ge}
\author[affiliation={1}]{Zikang}{Huang}
\author[affiliation={2}]{Yizhou}{Peng}
\author[affiliation={1}]{Jin}{Li}
\author[affiliation={1}]{Yuheng}{Lu}
\author[affiliation={1}]{Yu}{Jiang}
\author[affiliation={5}]{Nyima}{Tashi}
\author[affiliation={1,3},correspondingauthor]{Longbiao}{Wang}
\author[affiliation={4}]{Jianwu}{Dang}

\address{
    $^1$ Tianjin Key Laboratory of Cognitive Computing and Application, Tianjin University, China \\
    $^2$ Nanyang Technological University, Singapore \\
    $^3$ Huiyan Technology Company, Ltd., China \\
    $^4$ Chinese Academy of Sciences, China \\
    $^5$ Tibet University, China
}

\email{xc2024\_li@tju.edu.cn}


\usepackage{comment}
\usepackage{multirow}

\begin{document}

\maketitle

\begin{abstract}
    Speech Large Language Models have achieved breakthroughs in multilingual speech-to-text translation. However, existing approaches often overlook semantic commonalities across source languages, leading to biased translation performance. In this work, we propose POTSA (Parallel Optimal Transport for Speech Alignment), a new framework based on cross-lingual parallel speech pairs and Optimal Transport, designed to bridge high- and low-resource translation gaps. First, we introduce a Bias Compensation module to coarsely align initial speech representations. Second, we impose token-level OT constraints on a Q-Former using parallel pairs to establish fine-grained representation consistency. Then, we apply a layer scheduling strategy to focus OT constraints on semantically beneficial layers. Experiments on FLEURS show our method achieves SOTA performance, with +1.29 BLEU over five common languages and +2.93 BLEU on zero-shot languages, using only 10 hours of parallel speech per language
\end{abstract}

\section{Introduction}
The rapid advances of large language models (LLMs) \cite{gpt3,llama,qwen} in  natural language processing tasks \cite{gpt4} are driving a paradigm shift within the speech community \cite{survey}.
SpeechLLMs have consequently emerged, marking a transition from the conventional encoder-decoder architecture to a unified, language-model-centric framework \cite{salmon,qwenaudio,qwenomni}.
By mapping speech representation into a textual embedding space, these models support reasoning and generation directly from speech \cite{survey}.
Building on this foundation, Speech-to-Text Translation (S2TT) has emerged as one of the most promising applications of SpeechLLMs \cite{qwenaudio,stllm}.
The primary objective of S2TT is to efficiently convert speech input in a source language into textual output in a target language without intermediate transcription \cite{stsurvey}. This capability is particularly important in real-world multilingual applications \cite{stsurvey}, where a practical S2TT system is expected to generalize across diverse language pairs.
\begin{figure}
    \centering
    \includegraphics[width=0.9\linewidth]{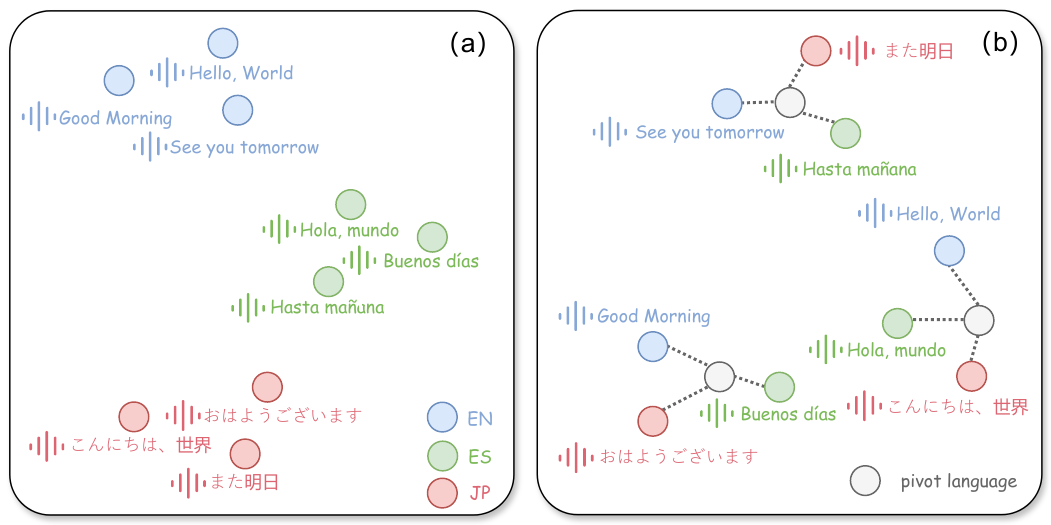}
    \caption{(a) Language-specific clusters before alignment.
    (b) Utterances with the identical semantics converge toward a shared representation (pivot language).}
      \label{fig:1}
    \vspace{-20pt}
\end{figure}
\begin{figure*}
        \centering
        \includegraphics[width=1\textwidth]{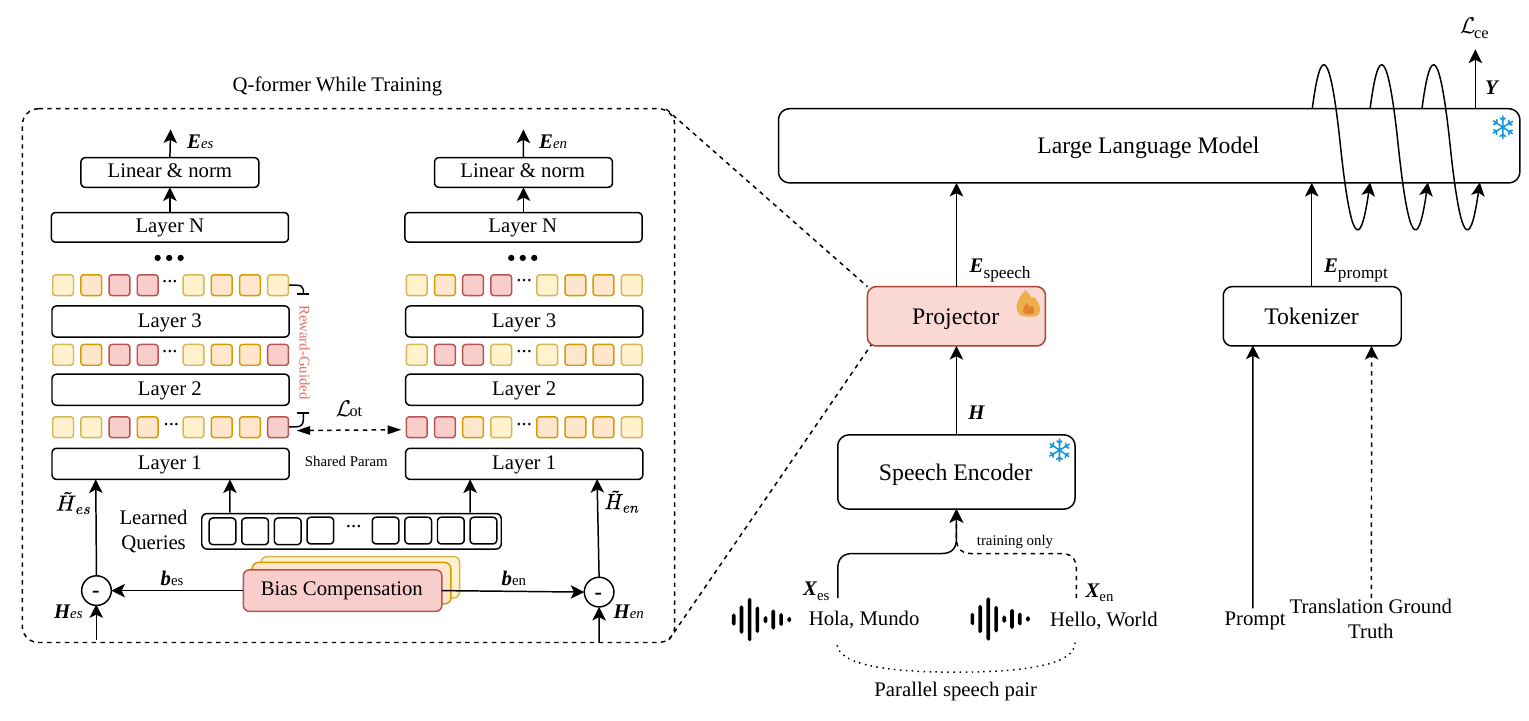}
        \caption{Model Architecture of Our Cross-Lingual Speech Alignment Framework}
    \label{fig:2}
    \vspace{-9pt}
\end{figure*}

However, meeting this requirement remains challenging: state-of-the-art S2TT models achieve high accuracy on high-resource language pairs but lag behind on low-resource ones \cite{stsurvey}. For example, English$\rightarrow$Chinese translation is typically much stronger than Japanese$\rightarrow$Chinese translation.
Although this performance bias is partially attributable to skewed training data distributions, the deeper underlying cause lies in insufficient cross-lingual comparability of speech representations \cite{onlyshared,whispershared}.
As illustrated in Figure \ref{fig:1}(a), we select the state-of-the-art multilingual encoder Whisper \cite{whisper} and visualize its speech representations, which form language-specific clusters:
samples from the identical language coalesce rather than being grouped by shared semantic content. This prevents the decoder from effectively reusing learned translation mappings, thus amplifying the performance gap.
Cross-lingual alignment is therefore the key to mitigating this bias. As illustrated in Fig \ref{fig:1}(b), this alignment encourages the model to induce a cross-lingually shared representation: an abstract 'pivot language'.
Nevertheless, current alignment research in the speech domain is confined to speech-to-text mappings.
By bridging the modality gap between speech and text, these methods facilitate knowledge transfer from machine translation (MT) decoders \cite{s2talign1,s2talign2,s2talign3}.
While such a strategy highlights the importance of cross-modal modeling, it implicitly carries two limitations: 
First, alignment is typically constrained to the speech-to-text direction, where the learning objective is defined by textual supervision \cite{s2talign2,s2talign3}. As a result, the speech encoder is forced to follow text-driven constraints, often compressing fine-grained acoustic details not represented in text.
Second, this paradigm treats each source language independently, neglecting the underlying transferable representations shared across languages \cite{multilingual}. This isolation restricts cross-lingual knowledge transfer and reduces the capacity for generalization.

Motivated by these observations, we propose \textbf{POTSA} (Parallel Optimal Transport for Speech Alignment), a novel alignment framework based on cross-lingual parallel speech pairs and Optimal Transport (OT) \cite{ot} to explicitly strengthen the consistency of cross-lingual representations for SpeechLLM in S2TT.
The framework trains only the Q-Former \cite{blip} to project encoder outputs cross-lingually, while freezing both the encoder and LLM.
First, we introduce a Bias Compensation module to mitigate language-specific shifts at the encoder output, narrowing cross-lingual gaps and preparing for fine-grained alignment within the speech representations.
Subsequently, we leverage cross-lingual parallel speech pairs with identical semantic content and impose token-level OT constraints within the Q-Former. This constraint achieves a finer-grained alignment of source speech representations.
Finally, we adopt layer-wise contributions and an online reward-guided scheduling strategy that applies OT constraints to the most effective layers.
Experiments on the FLEURS dataset demonstrate state-of-the-art performance, improving BLEU by +1.29 on average over five common languages and by +2.93 on zero-shot languages, using only 10 hours of parallel speech per source language for training. Our code is publicly available at \url{https://anonymous.4open.science/r/anonym1C8J}.

\section{Related Work}
\subsection{Representation Alignment}
Existing work on representation alignment in the speech domain predominantly focuses on speech-to-text mapping. Early studies pioneered unsupervised cross-modal embedding alignment \cite{s2talign1}, while subsequent approaches advanced this line of research through word-aligned contrastive learning \cite{s2talign2} and adaptive inner alignment for LLM integration \cite{s2talign3}. However, these text-driven, unidirectional paradigms inherently compress fine-grained acoustic details and overlook language-agnostic shared representations, thereby severely limiting cross-lingual knowledge transfer \cite{multilingual}.
We formulate our alignment objective using Optimal Transport (OT) \cite{ot} over source speech representations. Unlike strict point-to-point metrics (e.g., MSE or cosine similarity), which can fail when speech token indices are not aligned \cite{le2023pretraining}, OT computes a globally optimal soft matching under unknown correspondences, yielding stable, structure-preserving alignment signals \cite{xu2025optimaltransportregularizationspeech}.

\section{Proposed Method}

Figure \ref{fig:2} provides an overview of the SpeechLLM architecture adopted in this study. Our goal is to improve cross-lingual consistency of source speech representations in S2TT. We integrate an alignment framework into the model, as highlighted by the dashed box.
\subsection{Bias Compensation Module}
We propose a Bias Compensation Module to coarsely align source speech representations and enforce cross-lingual consistency.
We assume that each encoder output can be additively decomposed into a language-neutral component and a language-specific bias term \cite{bc1,bc2}, i.e., $H^{(i)}_x = \tilde{H}^{(i)}_x + b_x$.
Here, $\mathbf{H}_x^{(i)}$ denotes the encoder representation of the $i$-th utterance in language $x$, whereas $\mathbf{b}_{x}$ captures language-dependent bias shared by all utterances of that language.
In practice, we estimate $\mathbf{b}_{x}$ by averaging sentence-level representations after temporal pooling:
\begin{equation}
\mathbf{b}_{x} =
\frac{1}{N_{x}}
\sum_{n=1}^{N_{x}}
\text{Pool}\bigl(
\mathbf{H}_x^{(n)}
\bigr),
\end{equation}
where $N_x$ is the number of utterances in language $x$.Then, subtract this estimated bias from each utterance:
\begin{equation}
\tilde{\mathbf{H}}_x^{(n)} =
\mathbf{H}_x^{(n)}
- \mathbf{b}_{x}.
\end{equation}
During inference, $b_x$ is directly inherited from the training configuration.
This coarse alignment step reduces global language-specific discrepancies and establishes a more comparable representation space, thereby facilitating the subsequent fine-grained token-level alignment.
\subsection{Token-Wise OT Constraints for Q-Former}
After removing language-specific global bias via coarse alignment, we perform fine-grained alignment using Optimal Transport (OT) on cross-lingual parallel speech pairs.
Because each parallel pair carries identical semantics, we use these pairs as alignment anchors during training; in practice, only a limited amount is required to learn robust cross-lingual representations.
For each parallel pair, the final projection outputs are two token sequences, $E_u=\{u_i\}_{i=1}^n$ and $E_v=\{v_j\}_{j=1}^n$. For each selected Q-Former layer $\ell\in\mathcal{I}$, we also extract intermediate embeddings $E_u^{(\ell)}$ and $E_v^{(\ell)}$ to impose layer-wise alignment.
To enforce cross-lingual semantic consistency, we minimize the Sinkhorn distance—an entropy-regularized \cite{ot} OT objective. Let $Z\in\mathbb{R}_+^{n\times n}$ denote the transport plan under soft marginal constraints (typically uniform over valid tokens), and let $c(u_i,v_j)$ be the squared Euclidean ground cost. The per-layer loss is:
\begin{equation}
\mathcal{L}_{\text{ot}}^{(\ell)} = \min_{Z \geq 0} 
\sum_{i=1}^{n} \sum_{j=1}^{n} Z_{ij}\, c\left(u_i^{(\ell)}, v_j^{(\ell)}\right)
+ \varepsilon \sum_{i=1}^{n} \sum_{j=1}^{n} Z_{ij} \log Z_{ij},
\end{equation}
where $\varepsilon$ controls transport-plan smoothness. We compute this OT loss for every selected intermediate layer and average it in the total objective, which is jointly optimized with translation cross-entropy:
\begin{equation}
\mathcal{L}_{\text{total}} = \mathcal{L}_{\text{CE}} 
+ \frac{\alpha}{|\mathcal{I}|} \sum_{\ell \in \mathcal{I}} \mathcal{L}_{\text{ot}}^{(\ell)},
\end{equation}
where $\alpha$ balances the two terms, ensuring the model simultaneously learns accurate speech translation and cross-lingual token-level correspondence. 
The token-wise OT loss complements coarse bias compensation with fine-grained semantic alignment, improving cross-lingual consistency of source speech representations.
\subsection{Online Reward-Guided Layer Scheduling Strategy}
As aligning all layers may introduce redundancy and interfere with representation learning, we adaptively select the most effective layers for cross-lingual alignment in a deep architecture. We introduce an online, reward-guided layer scheduling strategy based on the Upper Confidence Bound (UCB) principle \cite{ucb}, combined with temperature-controlled softmax sampling.

For each candidate projection (Q-Former) layer $\ell\in\mathcal{I}$, we maintain:
(i) an exponentially weighted moving average (EMA) of its reward $Q_\ell$,
(ii) the selection count $n_\ell$, and
(iii) the task loss observed at its previous activation.
When layer $\ell$ is activated at iteration $t$, we compute a reward from the loss change and update the EMA:
\begin{equation}
Q_\ell^{(t)} = (1-\rho)\,Q_\ell^{(t-1)} + \rho\, r_\ell^{(t)},\quad \rho\in(0,1].
\end{equation}
We then compute the UCB utility and the temperature-controlled sampling distribution:
\begin{equation}
u_\ell^{(t)} = Q_\ell^{(t)} + \beta \sqrt{\frac{\log t}{\max(1,n_\ell^{(t)})}},
\end{equation}
\begin{equation}
p_{\ell}^{(t)} = \frac{\exp\left(u_{\ell}^{(t)}/\tau\right)}{\sum_{k \in \mathcal{I}} \exp\left(u_{k}^{(t)}/\tau\right)}.
\end{equation}
Here, $\beta>0$ controls exploration, and $\tau>0$ controls sampling sharpness.
This strategy concentrates cross-lingual alignment on the most informative layers, thereby improving cross-lingual consistency, efficiency, and stability.
\section{Experimental}
\subsection{Datasets and Training Setup}
We first trained a SpeechLLM on the large-scale CoVoST2 dataset \cite{covost2} using 364 hours of training data. The training proceeded in two stages: first, the model was pretrained on automatic speech recognition (ASR; English→English), followed by training on speech-to-text translation (S2TT; English→Chinese) tasks.
Subsequently, we performed multilingual mixed-supervision fine-tuning on FLEURS \cite{fleurs}, which covers over 100 languages with approximately 10 hours of supervised speech each. Five source languages—English (en), Japanese (ja), Spanish (es), Korean (ko), and Russian (ru)—were used to construct x→Chinese S2TT tasks.
We built a SpeechLLM within the SLAM-LLM framework \cite{slamasr}, where the speech encoder and language model are frozen, and cross-lingual alignment is achieved via a Q-Former (8 Transformer blocks, 80 query tokens) projection layer \cite{cost}. Training follows a two-stage schedule with learning rates of $1\times 10^{-4}$ and $1\times 10^{-5}$, respectively, with a warm-up mechanism to stabilize early training. The OT-based joint loss employs a weighting coefficient of 10. All experiments were conducted on NVIDIA RTX 4090 GPUs.
\subsection{Experimental Results}
\begin{table*}[ht]
        \caption{BLEU scores on the FLEURS test set for five training languages and two zero-shot evaluation languages}
  \centering
  \begin{tabular}{l|cccccc|c}
    \toprule
    \multirow{3}{*}{\textbf{Methods}} 
      & \multicolumn{6}{c|}{\textbf{FLEURS}} 
      & \multicolumn{1}{c}{\textbf{FLEURS-Zero}} \\
    \cmidrule(lr){2-7} \cmidrule(lr){8-8}
      & en$\rightarrow$zh & ja$\rightarrow$zh & es$\rightarrow$zh & ko$\rightarrow$zh & ru$\rightarrow$zh & avg.
      & avg. \\
    \midrule
    \multicolumn{8}{c}{\textbf{Baseline Models}} \\
    Qwen2-Audio \cite{qwen2audio} & 37.45 & 20.02 & 28.69 & 25.69 & 29.01 & 28.17 & 16.67 \\
    Qwen2.5-Omni \cite{qwenomni} & 38.35 & 23.76 & 27.81 & 27.05 & 29.69 & 29.33 & 16.91 \\
    MinMo \cite{minmo}        & 40.68 & 25.14 & \textbf{32.05} & 23.39 & \textbf{34.34} & 31.12 & - \\
    WhisperV3+Qwen2.5 (end-to-end)  & 40.09 & 24.32 & 30.14 & 27.07 & 31.13 & 30.55 & 17.91 \\
    \midrule
    \multicolumn{8}{c}{\textbf{Our Models}} \\
    Our Model & \textbf{40.87} & \textbf{25.97} & 31.10 & \textbf{28.97} & 32.30 & \textbf{31.84} & \textbf{20.84} \\
    \ \ (a) w/o Bias Comp. & 40.44 & 24.83 & 30.91 & 28.06 & 31.31 & 31.11 & 19.34 \\
    \ \ (b) w/o OT Align. & 40.22 & 25.22 & 30.89 & 25.91 & 31.40 & 30.73 & 18.51 \\
    \ \ (c) En-Anchor (Fixed) & 40.15 & 24.51 & 30.68 & 27.17 & 31.51 & 30.80 & 18.37 \\
    \ \ (d) En-Anchor (Train.) & 23.69 & 25.22 & 19.50 & 27.09 & 31.21 &  25.34 & 11.53 \\
                                   
    \bottomrule
  \end{tabular}
  \label{tab:fleurs_seen_unseen}
  \vspace{-10pt}
\end{table*}

Table \ref{tab:fleurs_seen_unseen} reports the BLEU scores on the FLEURS test set for five training languages and six zero-shot languages (French, German, Kabuverdianu, Cebuano, Asturian, Kyrgyz). We built a baseline with Whisper-v3 \cite{whisper}, a Q-Former, and Qwen-2.5-7B \cite{qwen}, and compared it against our model with alignment framework under the same training data. We also benchmark our system against the strongest recent SpeechLLMs.
The proposed method yields consistent BLEU improvements (average +1.29). The gains are particularly large for zero-shot languages (average +2.93). Although large-scale models like Minmo \cite{minmo}  (not publicly available) are trained on much larger corpora, our system already surpasses them on several metrics. As the alignment mechanism is model-agnostic, it can be integrated into larger systems to further enhance multilingual performance.

We conducted two types of ablation studies under an identical model backbone.
First, we evaluated the two alignment mechanisms separately: (a) the Bias Compensation module and (b) OT alignment. Table \ref{tab:fleurs_seen_unseen} shows that coarse alignment yields noticeable performance gains, while fine-grained OT alignment further improves overall translation quality.
Second, we studied how to construct parallel speech pairs for alignment. Our method adopts random pairwise alignment, where two randomly chosen languages are aligned symmetrically at each step.
We compare this choice with two English-anchor baselines: (c) frozen English, where the alignment loss is not back-propagated through the English branch; and (d) trainable English, where the alignment loss is back-propagated through the English branch.
As shown in Table \ref{tab:fleurs_seen_unseen}, the trainable-English baseline (d) yields the poorest performance, while frozen English (c) is slightly better. We attribute this to the English-centric asymmetry shared by both baselines: bidirectional updates in (d) can amplify representation drift and forgetting in low-resource languages, whereas (c) is more stable but still constrained by a fixed anchor. By contrast, our random pairwise alignment consistently achieves the best results by removing anchor dominance and fostering cross-lingual co-learning.

\begin{figure}
    \centering
    \includegraphics[width=1\linewidth]
    {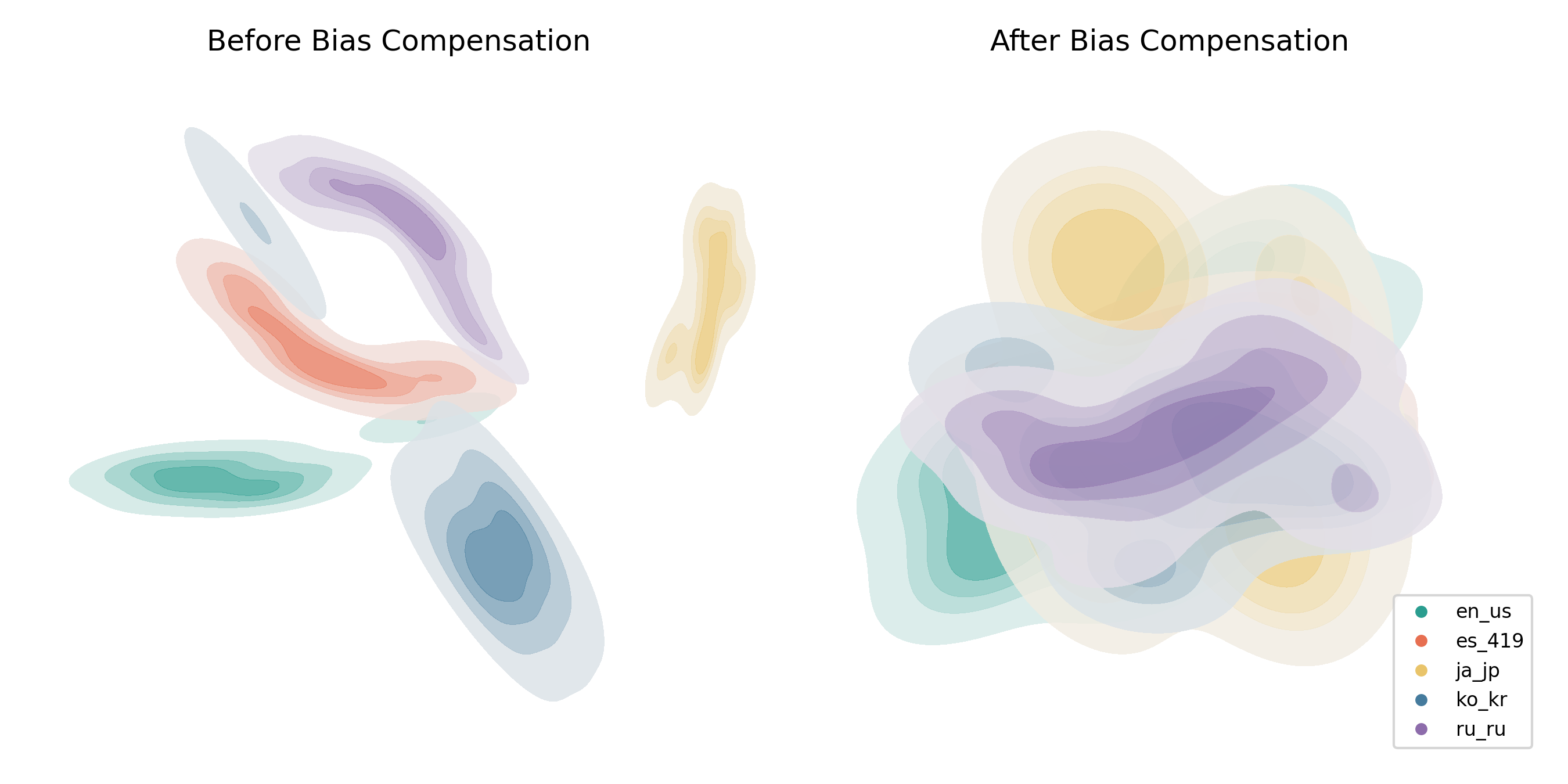}
    \includegraphics[width=1\linewidth]{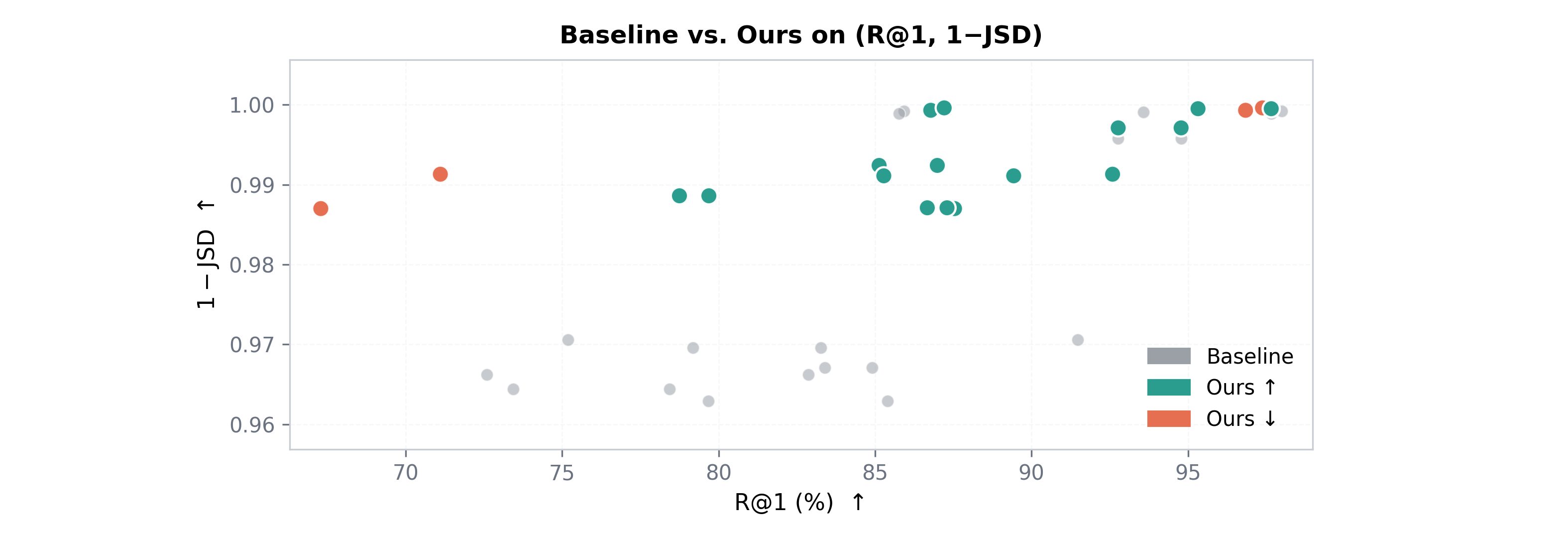}
    \caption{Alignment analysis: coarse bias compensation and fine-grained OT alignment improve cross-lingual consistency (R@1 and $1-\mathrm{JSD}$) and translation quality.}
    \label{fig:3}
\end{figure}
\begin{table}[th]
  \centering
  \caption{Comparison of alignment methods. Scores are the average BLEU with 95\% confidence intervals.}
  \begin{tabular}{ l c }
    \toprule
    \textbf{Alignment Methods} & \textbf{Avg. BLEU (95\% CI)} \\
    \midrule
    MSE                     & 30.52 $\pm$ 0.17 \\
    Cosine Similarity       & 30.74 $\pm$ 0.12 \\
    OT                 & \textbf{31.84 $\pm$ 0.09} \\
    \bottomrule
  \end{tabular}
  \label{tab:t3}
  
\end{table}
We evaluate whether our alignment modules bridge the interlingual gap, a prerequisite for robust zero-shot translation. 
Figure \ref{fig:3}(up) shows that Bias Compensation provides coarse-grained alignment by minimizing systematic interlingual offsets. Building on this coarse alignment, our OT-based alignment further encourages token-level speech representations consistency across languages.
To quantify these effects, we employ Recall@1 (R@1) to measure point-wise cross-lingual retrieval accuracy and 1-JSD to evaluate global distribution consistency. 
As shown in Figure \ref{fig:3}(down), language pairs migrate significantly toward the upper-right quadrant compared to the baseline. Crucially, this spatial alignment in both R@1 and 1-JSD directly translates into a +1.29 BLEU gain in translation performance. By unifying speech features into a language-agnostic space, the decoder can extract semantics without being confounded by interlingual variance, thereby minimizing translation errors caused by feature mismatch.

\begin{table}[th]
  \centering
  \caption{Performance comparison of different alignment strategies. Scores represent the average BLEU and their 95\% confidence intervals (CI). $^\dagger$ indicates statistical significance compared to the baseline ($p < 0.05$).}
  \begin{tabular}{ l l c }
    \toprule
    \textbf{ID} & \textbf{Layer selection strategy} & \textbf{Avg. BLEU (95\% CI)} \\
    \midrule
    (i)   & single                  & 31.36 $\pm$ 0.12 \\
    (ii)  & multi                   & 30.19 $\pm$ 0.15 \\
    (iii) & random                  & 31.50 $\pm$ 0.11 \\
    (iv)  & ours (across all layers)& 30.86 $\pm$ 0.14 \\
    (v)   & ours           & \textbf{31.84 $\pm$ 0.09}$^\dagger$ \\
    \bottomrule
  \end{tabular}
  \label{tab:t2}
  \vspace{-10pt}
\end{table}

To validate our choice of alignment method, we compared Optimal Transport (OT) against strict point-to-point metrics. As shown in Table \ref{tab:t3}, OT achieves the highest average BLEU score (31.84±0.09), outperforming both MSE and Cosine Similarity. This empirically confirms that OT's soft matching effectively overcomes the limitations of unaligned speech token indices; consequently, we adopt OT as our final alignment method.

To examine how the choice of alignment layer affects cross-lingual alignment, we evaluated five strategies:
(i) selecting a single layer;
(ii) selecting multiple fixed layers;
(iii) selecting a random layer;
(iv) applying the online reward-guided layer scheduling strategy across all Q-Former layers; and
(v) restricting the same strategy to the lower Q-Former layers.
As shown in Table \ref{tab:t2}, permitting a moderate degree of layer exploration is beneficial. The performance gains for our proposed method (v) are statistically significant ($p<0.05$ with 95\% CI) compared to the best baseline (iii), confirming that the improvement is not due to random fluctuation. Notably, Strategy (iv), which explores the full depth of the Q-Former, performs the worst, whereas Strategy (v) yields the best alignment quality.
Among the three naïve baselines, Strategy (iii) surpasses Strategy (i). It is worth noting that Strategies (ii) and (iv), which apply alignment to multiple or deeper Q-Former layers, yield the poorest results. We attribute this to higher-level features being closer to the decoder and primarily updated by the cross-entropy (CE) loss. Adding cross-lingual alignment at the upper layers creates competing objectives: the CE loss promotes translation accuracy, while the alignment term forces high-level representations to converge across languages, leading to degraded performance.

\section{Conclusions}
\label{sec:CONCLUSIONS}
We propose a cross-lingual speech alignment framework to bridge the gap in multilingual speech-to-text translation.
Experiments on the FLEURS benchmark show consistent translation gains with only a small number of parallel speech pairs.
Such limited data suffice to enhance cross-lingual consistency and unlock low-resource translation capability in existing large models.

\newpage
\section{Generative AI Use Disclosure}
During the preparation of this manuscript, the authors used generative AI tools to polish the English language, improve readability, and assist with \LaTeX{} formatting. These tools were not used to generate any scientific claims, experimental results, or significant parts of the manuscript.
\bibliographystyle{IEEEtran}
\bibliography{mybib}

@misc{gpt3,
      title={Language Models are Few-Shot Learners}, 
      author={Tom B. Brown and Benjamin Mann and Nick Ryder and Melanie Subbiah and Jared Kaplan and Prafulla Dhariwal and et al.},
      year={2020},
      eprint={2005.14165},
      archivePrefix={arXiv},
      primaryClass={cs.CL},
      url={https://arxiv.org/abs/2005.14165}, 
}

@misc{gpt4,
      title={GPT-4 Technical Report}, 
      author={OpenAI and Josh Achiam and Steven Adler and Sandhini Agarwal and Lama Ahmad and Ilge Akkaya and et al.},
      year={2024},
      eprint={2303.08774},
      archivePrefix={arXiv},
      primaryClass={cs.CL},
      url={https://arxiv.org/abs/2303.08774}, 
}

@misc{minmo,
      title={MinMo: A Multimodal Large Language Model for Seamless Voice Interaction}, 
      author={Qian Chen, Yafeng Chen, Yanni Chen and et al.},
      year={2025},
      eprint={2501.06282},
      archivePrefix={arXiv},
      primaryClass={cs.CL},
      url={https://arxiv.org/abs/2501.06282}, 
}

@misc{llama,
      title={LLaMA: Open and Efficient Foundation Language Models}, 
      author={Hugo Touvron and Thibaut Lavril and Gautier Izacard and Xavier Martinet and Marie-Anne Lachaux and Timothée Lacroix and Baptiste Rozière and Naman Goyal and Eric Hambro and Faisal Azhar and Aurelien Rodriguez and Armand Joulin and Edouard Grave and Guillaume Lample},
      year={2023},
      eprint={2302.13971},
      archivePrefix={arXiv},
      primaryClass={cs.CL},
      url={https://arxiv.org/abs/2302.13971}, 
}

@misc{qwen,
      title={Qwen2.5 Technical Report}, 
    author={Qwen and An Yang and Baosong Yang and et al.},
      year={2025},
      url={https://arxiv.org/abs/2412.15115}, 
}

@misc{salmon,
      title={SALMONN: Towards Generic Hearing Abilities for Large Language Models}, 
      author={Changli Tang and Wenyi Yu and Guangzhi Sun and Xianzhao Chen and Tian Tan and Wei Li and Lu Lu and Zejun Ma and Chao Zhang},
      year={2024},
      eprint={2310.13289},
      archivePrefix={arXiv},
      primaryClass={cs.SD},
      url={https://arxiv.org/abs/2310.13289}, 
}

@misc{qwenaudio,
      title={Qwen-Audio: Advancing Universal Audio Understanding via Unified Large-Scale Audio-Language Models}, 
      author={Yunfei Chu and Jin Xu and Xiaohuan Zhou and Qian Yang and Shiliang Zhang and Zhijie Yan and Chang Zhou and Jingren Zhou},
      year={2023},
      eprint={2311.07919},
      archivePrefix={arXiv},
      primaryClass={eess.AS},
      url={https://arxiv.org/abs/2311.07919}, 
}

@misc{qwen2audio,
      title={Qwen2-Audio Technical Report}, 
      author={Yunfei Chu and Jin Xu and Qian Yang and Haojie Wei and Xipin Wei and Zhifang Guo and Yichong Leng and Yuanjun Lv and Jinzheng He and Junyang Lin and Chang Zhou and Jingren Zhou},
      year={2024},
      eprint={2407.10759},
      archivePrefix={arXiv},
      primaryClass={eess.AS},
      url={https://arxiv.org/abs/2407.10759}, 
}

@article{qwenomni,
  title={{Qwen2.5-Omni Technical Report}},
  author={Xu, Jin and Guo, Zhifang and He, Jinzheng and et al.},
  journal={arXiv preprint arXiv:2503.20215},
  year={2025}
}

@misc{stllm,
  title={{Speech Translation with Large Language Models: An Industrial Practice}},
  author={Huang, Zhichao and Ye, Rong and Ko, Tom and et al.},
  year={2023},
  eprint={2312.13585},
  archivePrefix={arXiv},
  primaryClass={cs.CL},
  url={https://arxiv.org/abs/2312.13585}
}

@misc{multilingual,
      title={Neighbors and relatives: How do speech embeddings reflect linguistic connections across the world?}, 
      author={Tuukka Törö and Antti Suni and Juraj Šimko},
      year={2025},
      eprint={2506.08564},
      archivePrefix={arXiv},
      primaryClass={cs.CL},
      url={https://arxiv.org/abs/2506.08564}, 
}

@misc{stsurvey,
  title={{End-to-End Speech-to-Text Translation: A Survey}},
  author={Sethiya, Nivedita and Maurya, Chandresh Kumar and et al.},
  year={2024},
  eprint={2312.01053},
  archivePrefix={arXiv},
  primaryClass={cs.CL},
  url={https://arxiv.org/abs/2312.01053}
}

@misc{survey,
  title={{A Survey on Speech Large Language Models for Understanding}},
  author={Peng, Jing and Wang, Yucheng and Li, Bohan and et al.},
  year={2025},
  eprint={2410.18908},
  archivePrefix={arXiv},
  primaryClass={eess.AS},
  url={https://arxiv.org/abs/2410.18908}
}

@inproceedings{onlyshared,
  title={{Wave to Interlingua: Analyzing Representations of Multilingual Speech Transformers for Spoken Language Translation}},
  author={Abdullah, Badr M. and Shaik, Mohammed Maqsood and Klakow, Dietrich and et al.},
  booktitle={Proc. Interspeech 2024},
  pages={362--366},
  year={2024}
}

@misc{whispershared,
  title={{Cross-Lingual Transfer Learning for Speech Translation}},
  author={Ma, Rao and Qian, Mengjie and Fathullah, Yassir and et al.},
  year={2025},
  eprint={2407.01130},
  archivePrefix={arXiv},
  primaryClass={cs.CL},
  url={https://arxiv.org/abs/2407.01130}
}

@misc{whisper,
      title={Robust Speech Recognition via Large-Scale Weak Supervision}, 
      author={Alec Radford and Jong Wook Kim and Tao Xu and Greg Brockman and Christine McLeavey and Ilya Sutskever},
      year={2022},
      eprint={2212.04356},
      archivePrefix={arXiv},
      primaryClass={eess.AS},
      url={https://arxiv.org/abs/2212.04356}, 
}

@misc{blip,
      title={BLIP-2: Bootstrapping Language-Image Pre-training with Frozen Image Encoders and Large Language Models}, 
      author={Junnan Li and Dongxu Li and Silvio Savarese and Steven Hoi},
      year={2023},
      eprint={2301.12597},
      archivePrefix={arXiv},
      primaryClass={cs.CV},
      url={https://arxiv.org/abs/2301.12597}, 
}

@misc{ot,
  title={{Computational Optimal Transport}},
  author={Peyr{\'e}, Gabriel and Cuturi, Marco and et al.},
  year={2020},
  eprint={1803.00567},
  archivePrefix={arXiv},
  primaryClass={stat.ML},
  url={https://arxiv.org/abs/1803.00567}
}

@misc{ucb,
  title={{Upper-Confidence-Bound Algorithms for Active Learning in Multi-Armed Bandits}},
  author={Carpentier, Alexandra and Lazaric, Alessandro and Ghavamzadeh, Mohammad and et al.},
  year={2015},
  eprint={1507.04523},
  archivePrefix={arXiv},
  primaryClass={cs.LG},
  url={https://arxiv.org/abs/1507.04523}
}

@misc{s2talign1,
  title={{Unsupervised Cross-Modal Alignment of Speech and Text Embedding Spaces}},
  author={Chung, Yu-An and Weng, Wei-Hung and Tong, Schrasing and et al.},
  year={2018},
  eprint={1805.07467},
  archivePrefix={arXiv},
  primaryClass={cs.CL},
  url={https://arxiv.org/abs/1805.07467}
}

@misc{s2talign3,
      title={Adaptive Inner Speech-Text Alignment for LLM-based Speech Translation}, 
      author={Henglyu Liu and Andong Chen and Kehai Chen and Xuefeng Bai and Meizhi Zhong and Yuan Qiu and Min Zhang},
      year={2025},
      eprint={2503.10211},
      archivePrefix={arXiv},
      primaryClass={cs.CL},
      url={https://arxiv.org/abs/2503.10211}, 
}

@misc{s2talign2,
  title={{WACO: Word-Aligned Contrastive Learning for Speech Translation}},
  author={Ouyang, Siqi and Ye, Rong and Li, Lei and et al.},
  year={2023},
  eprint={2212.09359},
  archivePrefix={arXiv},
  primaryClass={cs.CL},
  url={https://arxiv.org/abs/2212.09359}
}

@misc{covost2,
  title={{CoVoST 2 and Massively Multilingual Speech-to-Text Translation}},
  author={Wang, Changhan and Wu, Anne and Pino, Juan and et al.},
  year={2020},
  eprint={2007.10310},
  archivePrefix={arXiv},
  primaryClass={cs.CL},
  url={https://arxiv.org/abs/2007.10310}
}

@misc{fleurs,
  title={{FLEURS: Few-Shot Learning Evaluation of Universal Representations of Speech}},
  author={Conneau, Alexis and Ma, Min and Khanuja, Simran and et al.},
  year={2022},
  eprint={2205.12446},
  archivePrefix={arXiv},
  primaryClass={cs.CL},
  url={https://arxiv.org/abs/2205.12446}
}

@article{cost,
  title={{CoT-ST: Enhancing LLM-Based Speech Translation with Multimodal Chain-of-Thought}},
  author={Du, Yexing and Ma, Ziyang and Yang, Yifan and et al.},
  journal={arXiv preprint arXiv:2409.19510},
  year={2024}
}

@misc{bc1,
  title={{On the Language Neutrality of Pre-Trained Multilingual Representations}},
  author={Libovick{\'y}, Jind{\v{r}}ich and Rosa, Rudolf and Fraser, Alexander and et al.},
  year={2020},
  eprint={2004.05160},
  archivePrefix={arXiv},
  primaryClass={cs.CL},
  url={https://arxiv.org/abs/2004.05160}
}

@inproceedings{bc2,
  title={{Correlations Between Multilingual Language Model Geometry and Cross-Lingual Transfer Performance}},
  author={Shah, Cheril and Chandak, Yashashree and Mane, Atharv Mahesh and et al.},
  booktitle={Proceedings of the 2024 LREC-COLING},
  pages={4059--4066},
  year={2024},
  address={Torino, Italia},
  publisher={ELRA and ICCL},
  url={https://aclanthology.org/2024.lrec-main.361/}
}

@misc{le2023pretraining,
      title={Pre-training for Speech Translation: CTC Meets Optimal Transport}, 
      author={Phuong-Hang Le and Hongyu Gong and Changhan Wang and Juan Pino and Benjamin Lecouteux and Didier Schwab},
      year={2023},
      eprint={2301.11716},
      archivePrefix={arXiv},
      primaryClass={cs.CL},
      url={https://arxiv.org/abs/2301.11716}, 
}

@misc{xu2025optimaltransportregularizationspeech,
      title={Optimal Transport Regularization for Speech Text Alignment in Spoken Language Models}, 
      author={Wenze Xu and Chun Wang and Jiazhen Yu and Sheng Chen and Liang Gao and Weihong Deng},
      year={2025},
      eprint={2508.08131},
      archivePrefix={arXiv},
      primaryClass={cs.CL},
      url={https://arxiv.org/abs/2508.08131}, 
}

@misc{slamasr,
      title={An Embarrassingly Simple Approach for LLM with Strong ASR Capacity}, 
      author={Ziyang Ma and Guanrou Yang and Yifan Yang and Zhifu Gao and Jiaming Wang and Zhihao Du and Fan Yu and Qian Chen and Siqi Zheng and Shiliang Zhang and Xie Chen},
      year={2024},
      eprint={2402.08846},
      archivePrefix={arXiv},
      primaryClass={cs.CL},
      url={https://arxiv.org/abs/2402.08846}, 
}

\end{document}